\pdfoutput=1
\documentclass[sigconf]{acmart}

\AtBeginDocument{%
  \providecommand\BibTeX{{%
    \normalfont B\kern-0.5em{\scshape i\kern-0.25em b}\kern-0.8em\TeX}}}
\definecolor{background_gray}{gray}{0.9}

\copyrightyear{2022}
\acmYear{2022}
\setcopyright{rightsretained}
\acmConference[SIGSPATIAL '22]{The 30th International Conference on Advances in Geographic Information Systems}{November 1--4, 2022}{Seattle, WA, USA}
\acmBooktitle{The 30th International Conference on Advances in Geographic Information Systems (SIGSPATIAL '22), November 1--4, 2022, Seattle, WA, USA}
\acmDOI{10.1145/3557915.3560947}
\acmISBN{978-1-4503-9529-8/22/11}

\usepackage{algorithm}
\usepackage{algorithmic}
\usepackage{mathrsfs}
\usepackage{amsthm}

\usepackage{amsmath}
\usepackage{amsfonts}
\usepackage{multirow}
\usepackage{enumitem}
\usepackage{colortbl}
\usepackage{color}

\newcommand{\revise}[1]{{\color{black}#1}}

\usepackage{subcaption}
\usepackage{stmaryrd}
\usepackage{placeins}
\usepackage{flushend}   

\makeatletter
\newcommand{\printfnsymbol}[1]{%
  \textsuperscript{\@fnsymbol{#1}}%
}
\makeatother

\begin{document}

\title{Periodic Residual Learning for Crowd Flow Forecasting}


\author{Chengxin Wang}
\email{cwang@comp.nus.edu.sg}
\affiliation{%
  \institution{National University of Singapore}
\country{Singapore}
}

\author{Yuxuan Liang}
\email{yuxliang@comp.nus.edu.sg}
\affiliation{%
  \institution{National University of Singapore}
  \country{Singapore}
  }

\author{Gary Tan}
\email{gtan@comp.nus.edu.sg}
\affiliation{%
  \institution{National University of Singapore}
  \country{Singapore}
}


\renewcommand{\shortauthors}{Wang et al.}

\begin{abstract}

Crowd flow forecasting, which aims to predict the crowds entering or leaving certain regions, is a fundamental task in smart cities. One of the key properties of crowd flow data is periodicity: a pattern that occurs at regular time intervals, such as a weekly pattern. To capture such periodicity, existing studies either fuse the periodic hidden states into channels for networks to learn or apply extra periodic strategies to the network architecture. In this paper, we devise a novel periodic residual learning network (PRNet) for a better modeling of periodicity in crowd flow data. Unlike existing methods, PRNet frames the crowd flow forecasting as a periodic residual learning problem by modeling the variation between the inputs (the previous time period) and the outputs (the future time period). Compared to directly predicting crowd flows that are highly dynamic, learning 
more stationary deviation is much easier, which thus facilitates the model training. Besides, the learned variation enables the network to produce the residual between future conditions and its corresponding weekly observations at each time interval, and therefore contributes to substantially more accurate multi-step ahead predictions. Extensive experiments show that PRNet can be easily integrated into existing models to enhance their predictive performance. 

\end{abstract}




\maketitle
\section{Introduction}

Nowadays, the development of intelligent transportation systems has drawn increasing attention as the number of vehicles grows over the years.
The total number of motor vehicles has reached 273 million in the U.S.~\cite{US}, and \revise{6.08 million} in Beijing by 2018, respectively, and it has grown to 6570 thousand in Beijing by 2020~\cite{BJ}.
To manage citywide transportation more efficiently, crowd forecasting aims to divide a city into multiple regions, i.e., even grid cells, and generate future vehicles' in/out-flow for each region. 
It is a crucial task that facilitates a wide range of applications in urban areas, such as assisting transportation managers to alleviate the congestion~\cite{ZhangHXXDBZ021}, guiding carsharing companies to pre-allocate vehicles~\cite{STMGCN19}, and helping travelers' decision-making~\cite{li2018multi}.

\begin{figure}[t!]
\begin{center}
  \includegraphics[width=1.0\linewidth]{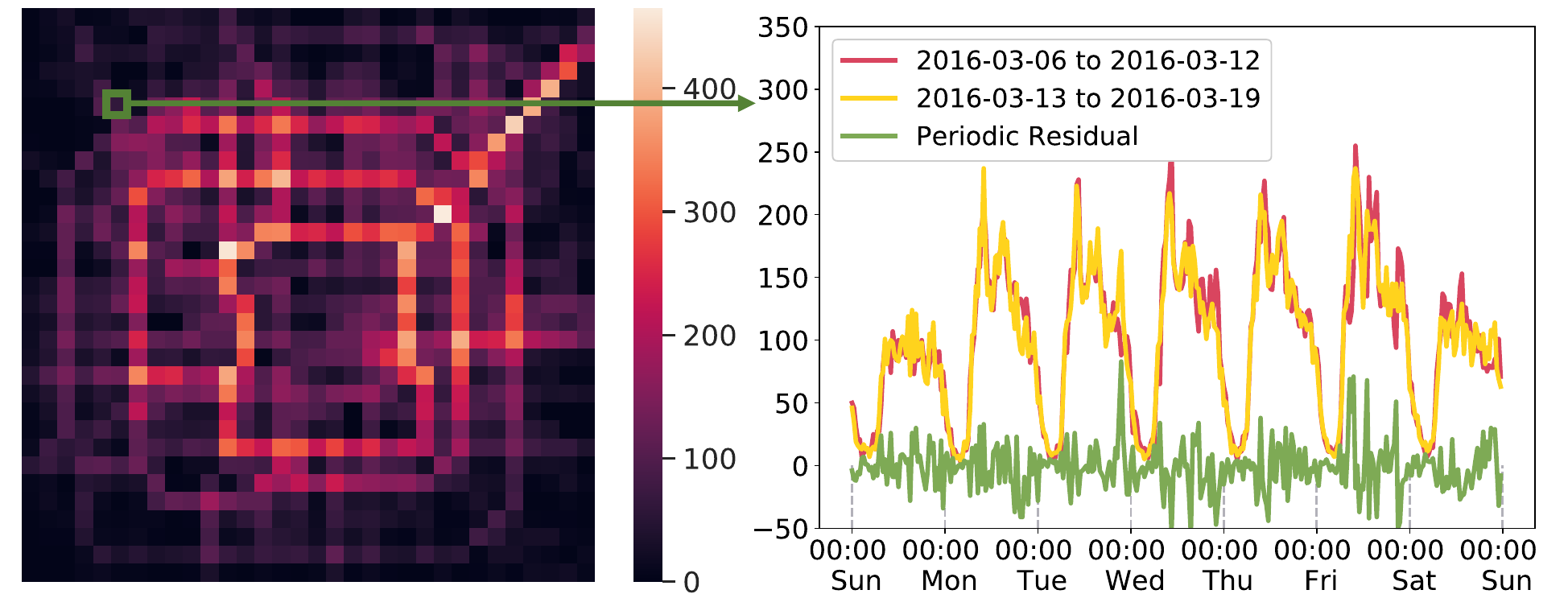}
\end{center}
\caption{
A visualization of crowd flows in Beijing. Left hand side: the city is divided into many regions; right hand side: the crowd inflow of a region during a period of two weeks, i.e., from 06 March 2016 to 19 March 2016.
}
\label{fig:flow_illustration}
\vspace{-1em}
\end{figure}

Spatio-temporal (ST) dependency~\cite{shekhar2015spatiotemporal,zhang2016dnn} is an important characteristic in crowd flow forecasting: one region's future crowd flow volume is conditioned on other regions' histories and its historical observations.
Mainstream works~\cite{zhang2017deep, lin2019deepstn+} employ convolutional neural networks (CNNs) to capture spatial correlations and utilize different sub-branches or channels to model temporal dependencies of different time scales.
Besides, there are some methods \revise{that adopt} recurrent neural networks (RNNs)~\cite{convlstm, zonoozi2018periodic, zhang2022urban} or Transformer~\cite{vaswani2017attention, xu2020spatial} to enhance temporal modeling via recurrent state transformations or attention mechanisms.
However, these models always require higher computational costs and larger storage compared to their CNN counterparts.
Meanwhile, more recent works~\cite{LiangOSWZZRZ21, liang2020revisiting} suggest that CNNs can effectively model the spatial and channel-wise correlations simultaneously with the Squeeze-and-Excitation (SE) mechanism~\cite{hu2018squeeze}.
With advanced mechanisms to express complex ST features, prior works have achieved promising prediction results.

\begin{figure}[t]
\begin{subfigure}{0.23\textwidth}
  \centering
    \includegraphics[height=2cm]{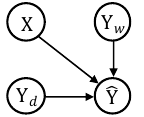}
  \caption{Existing periodic models}
\end{subfigure}
\begin{subfigure}{0.23\textwidth}
  \centering
  \includegraphics[height=2cm]{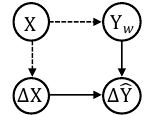}
  \caption{Ours}
\end{subfigure}
\caption{Graphical models for periodic \revise{modeling},
where $\mathbf{X}$ and $\hat{\mathbf{Y}}$ represent the \revise{current segment and the target segment}, respectively. $\mathbf{Y}_{d}$ and $\mathbf{Y}_{w}$ denote \revise{segments for the daily scale and the weekly scale}, respectively. 
The solid line indicates the direct relationship, and the dashed line denotes the indirect relationship.
}
\label{fig:graphical}
\vspace{-1em}
\end{figure}

\begin{figure}[!t]
\begin{subfigure}{0.18\textwidth}
  \centering
    \includegraphics[height=5.5cm]{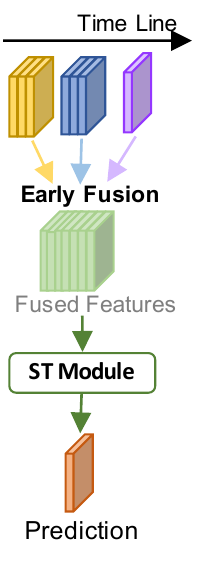}
  \caption{Feature-based model}
\end{subfigure}
\begin{subfigure}{0.25\textwidth}
  \centering
  \includegraphics[height=5.5cm]{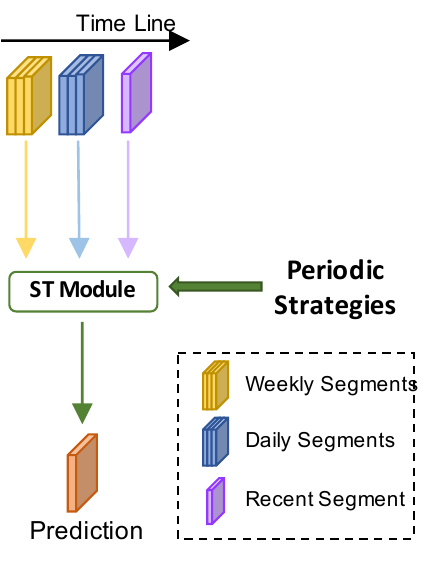} 
  \caption{Architecture-based model}
\end{subfigure}
\caption{Periodicity representation in ST neural networks.}
\vspace{-1em}
\label{fig:periodic}
\end{figure}

Another key characteristic in citywide crowd flow is periodicity~\cite{zhang2016dnn,yuan2017pred, shi2018discovering}.
As can be observed from Fig.~\ref{fig:flow_illustration}, crowd flow data show periodic patterns, e.g., daily and weekly.
For instance, on the daily scale, the volume in the \revise{region} follows a similar trend that increases during the morning and decreases during the night; on the weekly scale, the flow pattern trends to repeat every week (see the red and yellow line).
Existing works on \revise{representing} such periodic patterns can be summarized in Fig.~\ref{fig:graphical} (a). In detail, the multi-scale time intervals, such as the recent segment, daily segments, and/or weekly segments, are fed into the network for periodic learning.
These models can be grouped into two categories - \revise{\emph{feature-based} and \emph{architecture-based} models}. As shown in Fig.~\ref{fig:periodic} (a), the feature-based models view the multi-scale observations as different features and concatenate them as a tensor~\cite{lin2019deepstn+, liang2020revisiting} for the network to process.
However, the periodic information is mixed in the early stage, while being eliminated as the network depth increases.
To tackle this issue, the architecture-based models represent the periodicity more naturally via some extra periodic strategies (see Fig.~\ref{fig:periodic} (b)). 
For example, DeepST~\cite{zhang2016dnn} introduces different branches to capture the periodicity;
Periodic-CRN~\cite{zonoozi2018periodic} designs a loop-back mechanism to integrate the recurrent periodic representations; STDN~\cite{yao2019revisiting} utilizes the attention mechanism to calculate the similarity of ST representations between multi-scale segments.
However, these architectures inevitably induce high computational overheads and extra parameters, which may be prohibitive in large-scale crowd flow forecasting tasks.
Considering these facts, one may ask: \textit{can we address the periodic pattern in a more efficient manner?}

To answer this question, we first investigate the inherent periodic behavior of crowd data.
As shown in Fig.~\ref{fig:flow_illustration}, though the daily crowd flow often fluctuates, the volume difference of a certain region at the same time in successive weeks (we term it as \textit{periodic residual}) tends to be stable even in long-term trends (see the green line). 
As opposed to raw crowd flow data, periodic residual holds clearer patterns that are easier to learn~\cite{brockwell2016introduction}. We argue that periodic residual features are also consolidated representations extracted from raw data that can help to reduce the difficulties in modeling complex crowd flow patterns.
By learning such features, the network can be trained more efficiently, even with fewer parameters.
Based on this insight, we propose to think from a new perspective - introducing the residual concept to represent the periodic behavior.

In this paper, we present a novel architecture-based framework entitled \textbf{\underline{P}}eriodic \textbf{\underline{R}}esidual \textbf{\underline{Net}}work (PRNet) for multi-step ahead crowd flow forecasting.
Instead of designing complex ST extraction models or sophisticated periodic strategies, PRNet focuses on learning periodic residuals.
As depicted in Fig.~\ref{fig:graphical} (b), PRNet
converts the learning focus from directly generating predictions to computing the periodic residual. 
Formally, it structures a residual mapping that predicts the future temporal difference based on the past variation.
Then, the periodic learning structure in PRNet allows the network to: 
1) alleviate the computational costs by representing the periodicity with an efficient differencing function;
2) reduce memory consumption by learning the variation based on one periodic scale (e.g., weekly scale);
3) reduce redundant trainable parameters by encoding each periodic time interval into a shared parameter encoder;
4) make the network more effective and robust in long-term forecasting as the model generates predictions based on the learned periodical residual features at each time \revise{interval}.
To evaluate our periodic learning structure, we further integrate it into different baseline networks (e.g., DeepST \cite{zhang2016dnn}, ST-ResNet \cite{zhang2017deep}, and DeepLGR~\cite{liang2020revisiting}) and conduct extensive experiments on the real-world datasets.
Furthermore, we notice that \revise{existing works} are inefficient to capture the global ST correlations, and therefore introduce a lightweight ST enhanced network, named Spatial-Channel Enhanced (SCE) Encoder to jointly encode the most salient global spatial correlations as well as channel dependencies, i.e., \revise{ST representation. Our main contributions are summarized as follows:}

\begin{itemize}[leftmargin=*]

  \item We devise a simple yet effective periodic residual learning structure that learns the periodic residual at each time interval to improve the accuracy of multi-step ahead prediction. 
  This structure can be easily integrated into existing models.

  \item We introduce a lightweight Spatial-Channel Enhanced (SCE) Encoder to better capture global spatio-temporal dependencies.

  \item Experiments on two real-world datasets demonstrate that PRNet surpasses the state-of-the-art approaches in long-term predictions. We also show that the periodic residual learning structure brings significant improvements in performance for existing models, especially under the small data budget.
\end{itemize}

\begin{table*}[!htbp]
\centering
\tabcolsep=1.2 mm
\caption{The notations of crowd flow, where $P$ refers to the total number of selected periods and $p$ is the periodic index.}
\begin{tabular}{l|c|l|c|l}
\hline
Notation & Symbol & Definition & Color in Fig~\ref{fig:notation} & Shape\\
\hline
Closeness & $\mathbf{X}_{c}$ & Current segment & Purple & ${H \times W \times 2 \times T_{obs}}$\\
Periodic closeness & $\mathbf{X}_{p}$ & Periodic observations to the current segment & Blue &${H \times W  \times 2 \times T_{obs}}$\\
Prediction & $\mathbf{Y}$ & Target segment for prediction & Green & ${H \times W \times 2 \times T_{pred}}$\\
Periodic prediction & $\mathbf{Y}_{p}$ & Periodic observations to the target segment & Orange &$ {H \times W  \times 2 \times T_{pred}}$\\
Closeness residual & $\Delta\mathbf{X}$ & The residual between closeness and each periodic closeness & Pink &${P \times H \times W  \times 2 \times T_{obs}}$\\
Prediction residual & $\Delta\mathbf{Y}$ & The residual between prediction and each periodic prediction & Brown &${P \times H \times W  \times 2 \times T_{pred}}$\\
\hline
\end{tabular}
\label{notations}
\end{table*}

\section{Related Work}
\label{sec:relaed_work}

\noindent
\textbf{Grid-based Crowd Flow Forecasting}.
Crowd flow forecasting has been investigated for more than four decades. 
Early attempts employ statistical models~\cite{ahmed1979analysis, brockwell2016introduction,hoang2016fccf} to \revise{predict the condition of crowd flows}. 
In particular, some works~\cite{williams2003modeling,tran2015multiplicative} investigate the periodicity in crowd flows and apply the seasonal ARIMA to model it.
However, these \revise{works} rely on assumptions of linearity and stationarity and thereby cannot model the complex nonlinear ST dependency. 
Recently, deep learning models~\cite{zhang2017deep,STMGCN19,liang2020revisiting} have been used to capture the complex ST correlations. 
For example, DeepST~\cite{zhang2016dnn} and ST-ResNet~\cite{zhang2017deep} adopt CNN-based architectures to learn ST correlations and achieve higher \revise{forecasting} accuracy. 
Specifically, they integrate the periodicity into \revise{networks} by feeding multi-scale segments to different sub-branches.
For better periodic representations, other works \revise{model} the periodic pattern explicitly \revise{by} looping back the periodic representation dictionary~\cite{zonoozi2018periodic} or learning the temporal similarity~\cite{yao2019revisiting}.
However, they \revise{need} massive computation costs to \revise{loop back recurrent hidden states} or compute attention scores.
Recent efforts focus on improving spatial modeling for more accurate forecasts.
Graph neural networks (GNNs)~\cite{kipf2017semi,velivckovic2017graph} have become the frontier of spatial interactions learning in road-based \revise{networks}~\cite{zheng2020gman, jin2021hierarchical, han2021dynamic}, however, they have not demonstrated \revise{the} advantages over CNNs on \revise{grid-based problems.
Unlike the road-based network that is naturally a non-Euclidean graph, the even grid cells in the grid-based task are treated as pixels, without an explicit graph structure.} 
Meanwhile, CNNs have adequate ability to fully learn spatial interactions between grids via spatial kernels of each layer.
Recently, \revise{~\citeauthor{liang2020revisiting}~\cite{liang2020revisiting} shows} CNNs can effectively capture the ST correlations by 
jointly modeling spatial correlations and temporal dynamics. Besides grid-based crowd flow forecasting, there are some works on predicting on irregular regions \cite{sun2020predicting}.

\vspace{0.2em}
\noindent \textbf{CNNs and Attention Mechanisms}.
CNNs have been successfully applied to many domains, such as computer vision~\cite{he2016deep}, audio generation~\cite{oord2016wavenet}, 
crowd flow prediction~\cite{zhang2017deep}, etc.
Recent works~\cite{hu2018squeeze,hou2021coordinate} utilize gating and attention mechanisms to further enhance the feature interdependencies in CNNs.
Specifically, SENet~\cite{hu2018squeeze} introduces \revise{a squeeze-and-excitation (SE) operation} as the gating mechanism to recalibrate the channel-wise attention through the sigmoid function.
However, \revise{the global average pooling in SE suppresses spatial information, which makes the network fail to capture spatial correlations effectively.}
Although some works further introduce attention to enhance the spatial representation via operating additional convolutions layers on average- and max-pooled features~\cite{woo2018cbam} or \revise{employing} dilated convolutions to enlarge the receptive field~\cite{bamParkWLK18}, they fail to fully uncover the global correlations. 
DANet~\cite{fu2019dual} \revise{captures} global ST dependencies by extending the self-attention to position attention and channel attention.
However, it is computationally expensive since it takes all spatial information into account.
In this paper, we model the global ST representation in a computationally efficient manner by only considering the most salient features.

\begin{figure}[!b]
\begin{center}
  \includegraphics[width=1.0\linewidth]{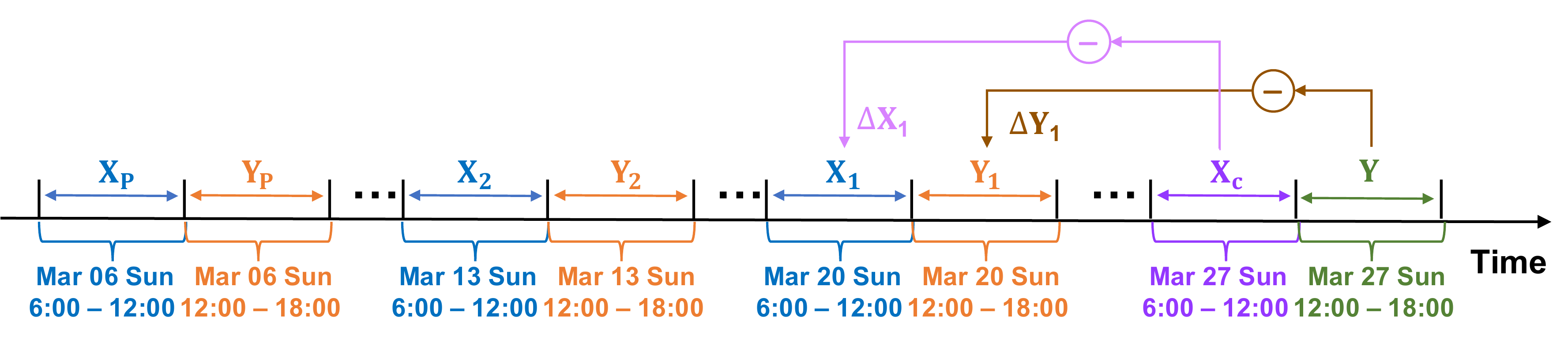}
\end{center}
\caption{
An example of the multi-scale segments notation under weekly scale, where the length of period $l$ is one week.
}
\label{fig:notation}
\end{figure}

\section{FORMULATION}
\label{sec:formulation}
In this section, we first define some notations and then formulate the problem of crowd flow forecasting.

\noindent \textbf{Definition 1 (Region)}: As shown in Fig.~\ref{fig:flow_illustration}, we follow \cite{zhang2017deep,Liang:2019,ouyang2020fine,li2020autost} to evenly partition an area of interest (such as a city) into $H \times W$ regions, \revise{i.e., grid cells}, based on their longitude and latitude.
\vspace{0.2em}

\noindent \textbf{Definition 2 (Crowd flow)}: The crowd flows at a certain time $\tau$ can be denoted as a 3D tensor $\mathcal{P}^\tau \in \mathbb{R}^{H \times W \times D}$, where $D$ is the number of attributes, e.g., inflow/outflow. Given a region $(h, w)$, \textit{inflow} refers to the total number of incoming traffic entering this region from other regions during a given time interval, while \textit{outflow} is the total number of outcoming traffic leaving this region.
\vspace{0.2em}

\noindent \textbf{Definition 3 (Closeness \& Periodic closeness)}: For better illustration, we define several segments in Table~\ref{notations} and Fig.~\ref{fig:notation}. Given the current timestamp $\tau$, the recent segment (i.e., closeness \cite{zhang2017deep}) and its corresponding periodic segments (i.e., periodic closeness in Table \ref{notations}) are denoted as:
\begin{equation}
\nonumber
\mathbf{X}_{c}= \mathcal{P}^{\tau-T_{obs}: \tau} = \left[\mathcal{P}^{\tau-T_{obs}}, \cdots, \mathcal{P}^{\tau} \right],
\end{equation}
\begin{equation}
\nonumber
\mathbf{X}_{1:P}= \left[\mathcal{P}^{t}_{1}\ , \mathcal{P}^{t}_{2}\, \cdots, \mathcal{P}^{t}_{P}\right]_{t=\tau -T_{obs}-l * p}^{\tau-l * p},
\end{equation}
where $T_{obs}$ is the length of recent observations, $P$ refers to the total number of selected periods, $l$ denotes the length of period, and $p$ is the period index. See more details in Table~\ref{notations} and Fig.~\ref{fig:notation}.
\vspace{0.2em}

\noindent \textbf{Definition 4 (Prediction \& Periodic prediction)}: After introducing closeness, we represent the target segment for prediction at time $\tau$ and its corresponding periodic segments as: 
\begin{equation}
\nonumber
\mathbf{Y} = \mathcal{P}^{\tau+1: \tau+T{pred}} = \left[\mathcal{P}^{\tau+1}, \cdots, \mathcal{P}^{\tau+T_{pred}} \right],
\end{equation}
\begin{equation}
\nonumber
\mathbf{Y}_{1:P}= \left[\mathcal{P}^{t}_{1}\ , \mathcal{P}^{t}_{2}\, \cdots, \mathcal{P}^{t}_{P}\right]_{t=\tau+1 -l * p}^{\tau + T_{pred} -l * p},
\end{equation}
where $T_{pred}$ is the length of target predictions.
\vspace{0.2em}

\noindent \textbf{Definition 5 (Closeness residual \& Prediction residual)}: \emph{Closeness residual} denotes the \revise{residuals} between $\mathbf{X}_{c}$ and $\mathbf{X}_{1:P}$, and \emph{prediction residual} represents the \revise{residuals} between ${\mathbf{Y}}$ and $\mathbf{Y}_{1:P}$ as:
\begin{equation}
\nonumber
\Delta{\mathbf{X}} = \left[\mathbf{X}_{c} - \mathcal{P}^{t}_{1}\, \cdots, \mathbf{X}_{c} - \mathcal{P}^{t}_{P}\right]_{t=\tau - T_{obs} -l * p}^{\tau -l * p},
\end{equation}
\begin{equation}
\nonumber
\Delta{\mathbf{Y}} = \left[\mathbf{Y} - \mathcal{P}^{t}_{1}\, \cdots, \mathbf{Y} - \mathcal{P}^{t}_{P}\right]_{t=\tau + 1 -l * p}^{\tau + T_{pred} -l * p},
\end{equation}

\begin{figure*}[ht]
\begin{center}
  \includegraphics[width=1.0\linewidth]{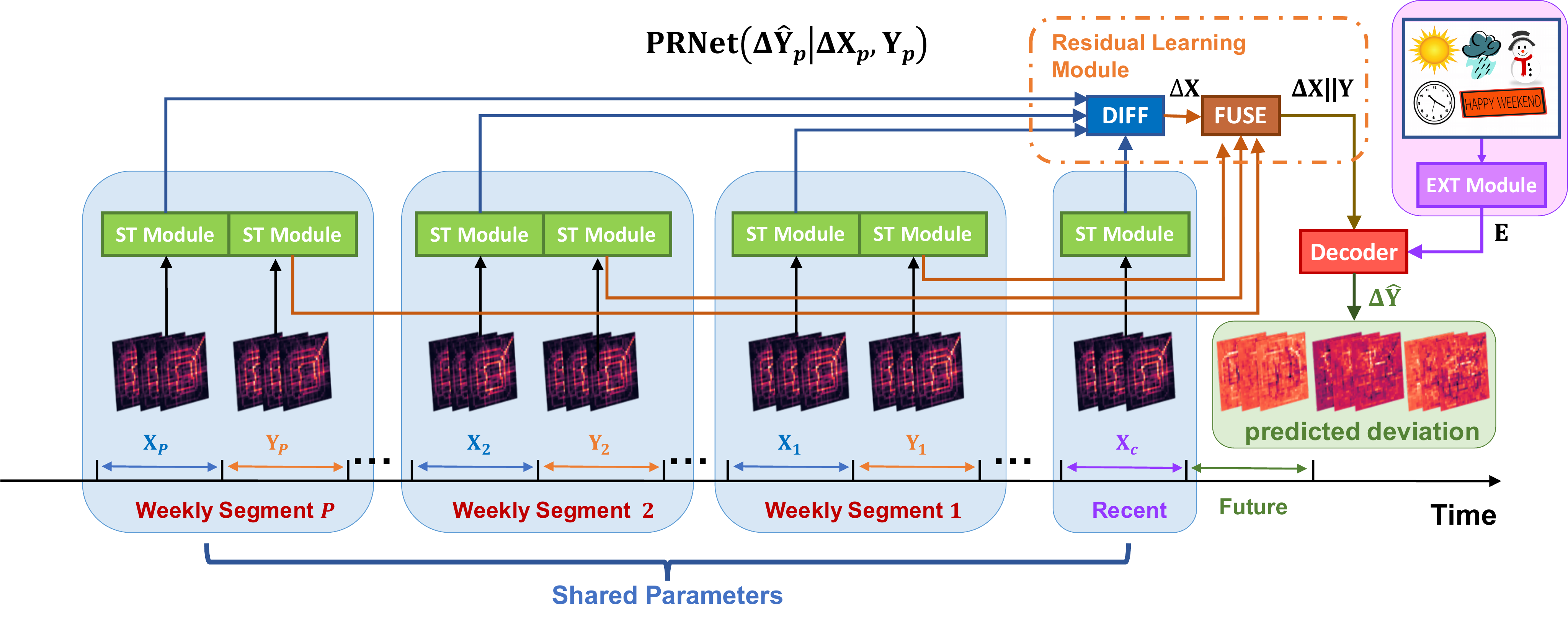}
\end{center}
\caption{
The overview of PRNet, where ST Module captures the ST correlations of each observed segment simultaneously. Then the network employs a differencing function (DIFF) to provide the closeness residual, and a fusion function (FUSE) to generate representations for the prediction residual. The decoder generates predicted deviations for all periodical weeks.
}
\label{fig:overview}
\end{figure*}

\vspace{0.2em}
\noindent
\textbf{Problem Statement (crowd flow forecasting): }
Given closeness $\mathbf{X}_{c}$, periodic closeness $\mathbf{X}_{1:P}$, periodic prediction $\mathbf{Y}_{1:P}$, the goal is to predict the prediction residual $\Delta\hat{\mathbf{Y}}$, which is equivalent to predict the future crowd flows $\hat{\mathbf{Y}}$.

\section{Periodic Residual Learning}
\label{sec:methodology}

Fig.~\ref{fig:overview} illustrates the pipeline of our proposed PRNet, whose core is a periodic residual learning structure. 
With the structure, PRNet reduces the data non-stationary by utilizing the closeness residual to assist prediction residual generation. 
For each segment (i.e., closeness, periodic closeness, and periodic prediction), we first fed the raw inputs to the shared ST Module for spatio-temporal representation.
Once we obtain the high-level features for each segment, we utilize a Residual Learning Module to learn prediction residual features. 
These features are then used to generate the predicted deviations via a Decoder. 
The details of PRNet will be elaborated in the following sections.

\subsection{Spatio-Temporal (ST) Module}
\label{stmodule}
Generality is one of the advantages of our proposed model.
Most of the existing Spatio-Temporal (ST) \revise{networks} can be easily integrated into PRNet as the Spatio-Temporal (ST) Module. 

\revise{A variety of ST networks has been designed to capture spatio-temporal dependencies.}
Based on the learning strategy, we group them into two categories, i.e., joint ST learning network and factorized ST learning network.
As its name suggests, joint ST learning networks simultaneously capture spatial and temporal dependencies by mapping the temporal inputs to CNN channels and utilizing the CNN kernels for spatio-temporal dependencies extraction~\cite{zhang2017deep, liang2020revisiting}.  In contrast, factorized ST learning networks decompose the modeling of ST into two separate dimensions, i.e., spatial dimension and temporal dimension. More specifically, they capture the spatial interactions and temporal dependencies sequentially via convolutional layers~\cite{yao2019revisiting} or convolutional graph layers~\cite{ZhangHXXDBZ021} for spatial dimension and recurrent mechanisms~\cite{convlstm} or attention mechanisms~\cite{ZhangHXXDBZ021} for temporal dimension. 
Among these two schemes, joint ST learning networks are usually applied to \revise{grid-based crowd flow forecasting for two reasons: 1) The grid cells in the tasks are even and can be treated as pixels.} 2) Recurrent and attention operations usually require high computation costs, especially when the multi-scale time intervals need to be considered~\cite{lin2019deepstn+}.

In PRNet, \revise{ST Module extracts high-level spatio-temporal representations (denoted as $\mathbf{h}$) for each segment via the ST network}:
\begin{equation}
{\mathbf{h}}= f(\mathcal{P}^{t:t+\revise{T_{obs}}}; \mathbf{W}_{st}),
\end{equation}
where $\mathbf{h} \in \mathbb{R}^{H \times W \times C}$ is the output features; $f$ represents the function of an ST network; $\mathbf{W}_{st}$ denotes the learnable parameters, and $t$ is the start timestep of a given time interval.
Unlike existing attempts that encode multi-scale time intervals into compacted features \cite{lin2019deepstn+,liang2020revisiting}, each segment in our PRNet is fed separately to a shared ST Module to save parameter usage.
More details of our proposed ST Module will be introduced in Section~\ref{scemodule}.

\begin{figure*}[ht]
\begin{center}
  \includegraphics[width=1\linewidth]{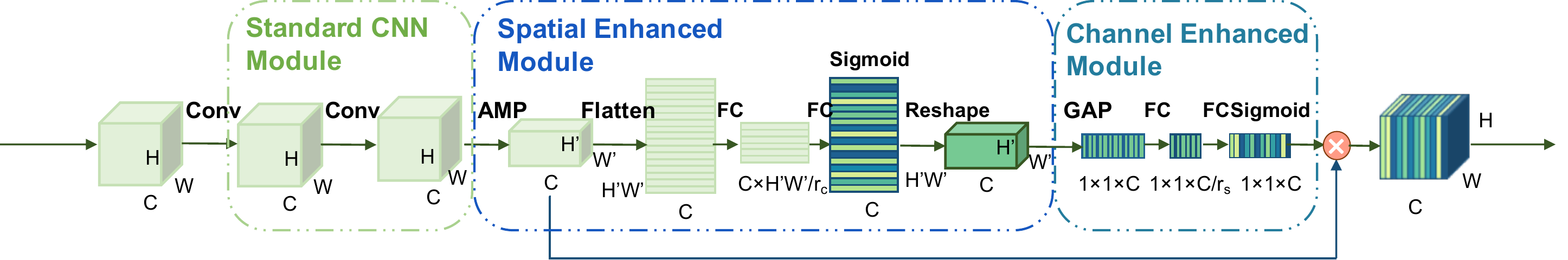}
\end{center}
\caption{
An illustration of Spatial-Channel Enhanced (SCE) Block in SCE Encoder. 
}
\label{fig:sceblock}
\end{figure*}

\subsection{Residual Learning Module}
Statistical methods~\cite{williams1999modeling, brockwell2016introduction} have demonstrated robust prediction by removing trends and seasonality given time series data. In light of these approaches, we introduce a similar concept to deep learning networks by devising a residual learning module to eliminate the sequential seasonality (i.e., periodicity in this paper). This new module aims to learn the high-level features of the periodic residual that are less complex but still maintain the periodic information. It consists of two functions: differencing function and fusion function.

\vspace{0.3em}
\noindent
\textbf{Differencing function (DIFF)} removes the seasonality and provides the periodic closeness residual as a reference of temporal shifting to the network. 
Traditional statistical approaches use the subtraction function to eliminate the seasonality~\cite{brockwell2016introduction}. Thus, we also choose it as our differencing operation since the learned ST features from the ST Module map to their corresponding raw observations.
Then the hidden states of periodic closeness residual can be calculated by subtracting the hidden state of closeness ${\mathbf{h}}_{x}$ from the hidden state of periodic closeness $\mathbf{h}_{px}$ generated by ST Module:

\begin{equation}
\nabla_{d}\mathcal{H} = {\mathbf{h}}_{x}- {\mathbf{h}}_{px},
\end{equation}
where $\nabla_{d}$ denotes the differencing operator, and $\nabla_{d}\mathcal{H} \in \mathbb{R}^{P \times H \times W \times C}$. Note that dimension broadcast is used.

\vspace{0.3em}
\noindent
\textbf{Fusion function (FUSE)} \revise{generates} the prediction residual features for the Decoder to produce the prediction residual (i.e., \revise{residuals} between future crowd flows $\mathbf{Y}$ and its corresponding periodic predictions $\mathbf{Y}_{p}$).
As the periodic closeness residual and the periodic predictions can provide the time-shifting references for the prediction residual,
we consider such information into the network by adopting a concatenation function followed by a canonical linear layer on their features:
\begin{equation}
\tilde{\mathcal{H}} = \mathbf{W}_{d}(\nabla_{d}\mathcal{H} \parallel {\mathbf{h}}_{py})
\end{equation}
where $\parallel$ is the concatenation operation, and $\mathbf{W}_{d}$ denotes learnable parameters.
Therefore, the embedded vector $\tilde{\mathcal{H}} \in \mathbb{R}^{P \times H \times W \times C}$ can represent the hidden states of the prediction residual, which are conditioned on the features of closeness residual and periodic prediction. 
It enables the model to learn \revise{the} deviations between future conditions and its historical observations.
It is worth noting that with the residual learning strategy, PRNet provides more stationary features to the network so that it increases the model capacity with no extra costs in parameter space.

\subsection{External Module \& Decoder}
\revise{External factors, such as date, event, and weather, can affect crowd flows~\cite{zhang2017deep, LiangOSWZZRZ21,zhou2021urban}.
The External (EXT) Module works on encoding these factors and outputs the external factor embedding $\mathbf{E}$.}
Same as the ST Module, the EXT Module in PRNet is also a general module, which can be plugged by any existing attempts~\cite{zhang2016dnn, zhang2017deep} or be omitted~\cite{liang2020revisiting}. 
The same form is for the Decoder. And the default Decoder of PRNet is a fully-connected layer \revise{(FC)}. However, instead of generating absolute values, PRNet focuses on fully uncovering the temporal shifting in periodicity by predicting the variation 
$\Delta \hat{\mathbf{{Y}}}$ between the future and its corresponding historical average flows based on \revise{$\tilde{\mathcal{H}}$ or the concatenation of $\tilde{\mathcal{H}}$ and $\mathbf{E}$.} 
To strengthen the robustness of our model, all $P$ historical segments are considered.
Therefore, we define the loss function as:
\begin{equation}
\mathcal{L}(\theta)=\sum\nolimits_{\tau=1}^{ T_{p r e d}}\left\|\Delta\hat{\mathbf{{Y}}}^{\tau}- \Delta\mathbf{Y^{\tau}}\right\|_{1},
\end{equation}
where $\theta$ denotes learnable parameters in the model. Then the predicted deviation $\Delta \hat{\mathbf{Y}} \in \mathbb{R}^{P \times H \times W \times 2 \times T_{pred}}$ \revise{can be easily converted} to the absolute crowd flows $\hat{\mathbf{Y}}\in \mathbb{R}^{H \times W \times 2 \times T_{pred}}$:
\begin{equation}
\hat{\mathbf{Y}} = \sum\nolimits_{i=1}^{P} (\Delta\hat{\mathbf{Y}} + \mathbf{Y}_{p}) / {P},
\end{equation}
where $P$ is the total number of the periodic segments. 

\section{ST Module - SCE Encoder
\label{scemodule}}

In Section~\ref{stmodule}, we have discussed that joint ST learning networks are widely used as the backbone to capture the spatio-temporal features of grid-based \revise{crowd flows}. 
Among them, many works~\cite{zhang2017deep,lin2019deepstn+} focus on modeling long-range spatial correlations. Because the one grid cell's crowd flow can be affected by distant neighbors due to vehicles moving very far given a large time interval.
To learn long-range spatial dependencies, existing works stack multiple CNN layers to larger the receptive field~\cite{zhang2016dnn, zhang2017deep} or apply fully-connected layers to aggregate the features from all grids~\cite{lin2019deepstn+}. However, they have underestimated the temporal relationship between channels within feature maps.
Recently, \citeauthor{liang2020revisiting}~\shortcite{liang2020revisiting} adopts squeeze-and-excitation networks (SENet)~\cite{hu2018squeeze} to explicitly model the channel-wise relations to enhance ST representation learning. 
However, it fails to capture complex global patterns as it squeezes global spatial features at each block.
Although it further introduces a pyramid CNN structure to abstract the spatial features at different levels for global spatial learning, it also requires additional computational costs.
To address the above issue, we propose a Spatial Channel Enhanced (SCE) Encoder as the ST Module in our proposed PRNet. It enhanced the SENet by introducing a lightweight global spatial enhanced module to emphasize the global salient spatial features. The SCE Encoder contains two main components: Embedding Layer and Spatial-Channel Enhanced Block.

\subsection{Embedding Layer}
We follow the previous studies \cite{zhang2017deep,lin2019deepstn+} to employ an embedding layer for a feature transformation. In detail, this layer converts each observed segment $\mathcal{P} \in \mathbb{R}^{ H \times W \times D \times T}$ to feature maps $\mathbf{z} \in  \mathbb{R}^{ H \times W \times C}$ through a convolutional operation with kernel size 1, where $T$ denotes the total time intervals of the segment, namely $T_{obs}$ for closeness and $T_{pred}$ for prediction. 

\subsection{Spatial-Channel Enhanced Block}
In Fig.~\ref{fig:sceblock}, we illustrate a single SCE block in SCE Encoder. It comprises three main modules: Standard CNN Module, Spatial Enhanced Module, and Channel Enhanced Module. 
Since the spatial and temporal information has been indexed to dimensions and channels, the Standard CNN Module can capture the local spatio-temporal correlation via convolution layers: 
\begin{equation}
\vec{\mathbf{h}}^{(m)}=\mathbf{W}_{f2}^{(m)} \star\left(\delta\left(\mathbf{W}_{f1}^{(m)} \star \mathbf{h}^{(m)}\right)+b_{f1}^{(m)}\right)+b_{f2}^{(m)}
\end{equation}
where $\mathbf{W}_{f1}$, $\mathbf{W}_{f2}$, $b_{f1}$ and $b_{f2}$ are learnable parameters, $\star$ refers to a convolution operator, $\delta(\cdot)$ is ReLU activation function, and $m$ denotes index number of SCE Blocks. Note that $\mathbf{h}^{(0)}$ is $\mathbf{z}$ and $\vec{\mathbf{h}} \in \mathbb{R}^{H \times W \times C}$. We will omit the index $m$ for the same block in the following sections.

\noindent
\textbf{Spatial Enhanced Module (SEM)} enhances the standard CNN by selecting the salient features globally for better spatial representation.
To achieve it, we adopt adaptive max pooling (AMP) to down-sample the hidden state $\vec{\mathbf{h}}$ by selecting most important features $\tilde{\mathcal{S}} \in \mathbb{R}^{H^{\prime} \times W^{\prime} \times C}$ and translate it to $\mathcal{S}^{\prime} \in \mathbb{R}^{C \times  H^{\prime} W^{\prime}}$.
Then the excitation operator~\cite{hu2018squeeze} is adopted to adaptively recalibrate these global salient features for better spatial correlation modelling:
\begin{equation}
\hat{\mathbf{h}}_{s}=\sigma(g(\mathcal{S}^{\prime}, \mathbf{W}_{s}))=\sigma\left( \delta\left( \mathcal{S}^{\prime}\mathbf{W}_{s1}\right)\mathbf{W}_{s2}\right),
\label{eq:excitation}
\end{equation}
where $\sigma$ refers to sigmoid function, $\delta$ denotes the ReLU function, 
$g(\cdot)$ represents the gated function,
$\hat{\mathbf{h}}_{s} \in \mathbb{R}^{C \times H^{\prime}W^{\prime}}$,
$\mathbf{W}_{s1} \in \mathbb{R}^{H^{\prime}W^{\prime} \times r_{s} }$,
$\mathbf{W}_{s2} \in \mathbb{R}^{r_{s} \times H^{\prime} W^{\prime}}$, and $r_{s} \ll H^{\prime} W^{\prime}$. 
By using learnable parameters $\mathbf{W}_{s1}$ and $\mathbf{W}_{s2}$ to reduce and increase the feature dimensions sequentially, the gated function enables the network to dynamically control the bypass signals and only capture the most salient features.
Then we reshape $\hat{\mathbf{h}}_{s}$ and obtain the final encoded global spatial feature $\widetilde{\mathbf{h}}_{s} \in \mathbb{R}^{C \times H^{\prime} \times W^{\prime}}$.

\noindent
\textbf{Channel Enhanced Module (CEM)} Except for spatial correlations, dynamic spatio-temporal dependencies need to be considered in crowd flow tasks.
We thus propose to use CEM to learn spatial correlations and temporal dependencies simultaneously for better ST understanding.
It first summarizes the global spatial features into a channel descriptor, then the descriptor captures the spatio-temporal correlations based on the channel dimension.
We adopt global average pooling (GAP) to squeeze the global spatial features and generate channel-wise statistics:
\begin{equation}
\mathbf{c}=\frac{1}{H^{\prime} \times W^{\prime}} \sum_{h=1}^{H^{\prime}} \sum_{w=1}^{W^{\prime}} \widetilde{\mathbf{h}}_{s}(h, w),
\end{equation}
where $\mathbf{c} \in \mathbb{R}^{C}$. Then a similar strategy as Eq.~\ref{eq:excitation} is used to enhance the spatio-temporal representation by producing the compacted channel-wise features:
\begin{equation}
\widetilde{\mathbf{h}}=\sigma(g(\mathbf{c}, \mathbf{W}_{c}))=\sigma\left(\mathbf{W}_{c2} \delta\left(\mathbf{W}_{c1} \mathbf{c}\right)\right)
\end{equation}
where $\mathbf{W}_{c1} \in \mathbb{R}^{\frac{C}{r_{c}} \times C}$, $\mathbf{W}_{c2} \in \mathbb{R}^{C \times \frac{C}{r_{c}}}$, $r_{c}$ is the reduction ratio, and $\widetilde{\mathbf{h}} \in \mathbb{R}^{1 \times 1 \times C}$. 
The final output of one SCE Block can be obtained by scaling the compacted features $\widetilde{\mathbf{h}}^{(m)} \in \mathbb{R}^{1 \times 1 \times C}$ and the ST feature map $\vec{\mathbf{h}}^{(m)}$:
\begin{equation}
\mathbf{h}^{(m+1)} = \widetilde{\mathbf{h}}^{(m)}\vec{\mathbf{h}}^{(m)}.
\end{equation}
By stacking multiple SCE blocks, SCE Encoder can model long-term spatio-temporal dependencies effectively. 
We stack a total number of $M$ SCE blocks in the SCE Encoder.
The receptive field of succeeding blocks in SCE Encoder is larger than the receptive field of former blocks. Therefore, our model constructs simple direct ST interactions between grids in former blocks and indirect global ST connections in the succeeding blocks. 
To this end, SCE Encoder can efficiently describe correlations between grids over time.

\section{Experiments}
\label{sec:experiments}

\subsection{Experimental Settings}
\subsubsection{Datasets} 
\label{sec:dataset}
We conduct experiments on two real-world datasets \cite{zhang2016dnn}, i.e., TaxiBJ and BikeNYC. 
TaxiBJ dataset is the crowd flow dataset, which is obtained through taxicab GPS data. It comprises four sub-datasets - P1, P2, P3, and P4. And BikeNYC dataset records the bike trajectory information which is extracted from the NYC bike system. The detailed statistical information of the datasets is described in Table~\ref{dataset}.
Besides, the external features of the datasets include holidays, weather conditions, temperature, and wind speed.

\begin{table}[!htbp]
\centering
\small
\tabcolsep=1.2mm
\caption{The statistic of TaxiBJ and BikeNYC dataset.}
\label{dataset}
\begin{tabular}{l|l|l|l|l}
\hline
\multirow{2}{*}{\textbf{Dataset}} & \multirow{2}{*}{\textbf{Grid Map}} & \textbf{Time Interval}  & \textbf{Time} & \textbf{Min - Max} \\
&&\textbf{(mm/dd/yyyy)}& \textbf{Span} & \textbf{Value}\\
\hline
TaxiBJ-P1& (32, 32)  & 07/01/2013 - 10/31/2013 & 30 mins & 0 - 1230 \\
TaxiBJ-P2& (32, 32) & 03/01/2014 - 06/30/2014  & 30 mins & 0 - 1292  \\
TaxiBJ-P3 & (32, 32) & 03/01/2015 - 06/30/2015 & 30 mins & 0 - 1274 \\
TaxiBJ-P4 & (32, 32)  & 11/01/2015 - 04/10/2016 & 30 mins & 0 - 1250  \\
\hline
BikeNYC& (16, 8)  & 04/01/2014 - 30/09/2014 & 60 mins & 0 - 267  \\
\hline
\end{tabular}
\end{table}

\begin{table}[!htbp]
\centering
\small
\tabcolsep=1.7mm
\caption{The details of data samples over two datasets.}
\vspace{-0.5em}
\begin{tabular}{l|c|c|c|c|c|c}
\hline
\multirow{2}{*}{\textbf{Dataset}} & \multirow{2}{*}{\textbf{Days}} 
& \textbf{Total} 
& \textbf{Missing} &
\multicolumn{3}{c}{\textbf{Sample Size}}\\\cline{5-7}
& &\textbf{Timeslots}& \textbf{Ratio} & \textbf{Train}
& \textbf{Valid}
& \textbf{Test}\\
\hline
TaxiBJ-P1& 121 & 5808 & 15.8\% & 2164 & 720 & 720 \\
TaxiBJ-P2 & 119 & 5712 & 16.3\% & 2080& 693 & 693 \\
TaxiBJ-P3 & 122 & 5596 & 4.4\% & 2665 & 887 & 887 \\
TaxiBJ-P4 & 162 & 7776 & 7.2\% & 3507 & 1168 & 1168\\
\hline
BikeNYC & 183 & 8784  & 50.0\% & 1101 & 366 & 366\\
\hline
\end{tabular}

\label{dataset_details}
\end{table}

\begin{table*}[ht]
\centering
\small
\tabcolsep=7 mm
\caption{
Model comparison on the TaxiBJ dataset in terms of performance and parameter size, where K denotes thousand and M denotes million. The format of numerical results is "mean $\pm$ standard deviation" (the lower results are better).
}
\label{quantitative_result}
\begin{tabular}{l|c|cc|cc}
\hline
\multirow{2}{*}{\textbf{Method}} & \multirow{2}{*}{\textbf{\# Params}}& \multicolumn{2}{c|}{\textbf{P1}} & \multicolumn{2}{c}{\textbf{P2}}\\\cline{3-6}
&& MAE &  RMSE  &  MAE  & RMSE  \\
\hline
HA &-& 16.91 & 31.49 
& 13.65 & 23.97  \\
DeepST & 380K & 15.68 $\pm$ 0.43 & 26.69 $\pm$ 0.79 &
15.61 $\pm$ 0.35 & 25.48 $\pm$ 0.56\\
ST-ResNet & 3077K & 13.84 $\pm$ 0.13 & 23.48 $\pm$ 0.16 
& 13.74 $\pm$ 0.42 & 22.87 $\pm$ 0.57\\
ConvLSTM & 1839K & 11.77 $\pm$ 0.06 & 20.19 $\pm$ 0.14 
& 12.47 $\pm$ 0.14 & 21.89 $\pm$ 0.36 \\
DeepSTN+ & 105M & 13.41 $\pm$ 0.28 & 25.51 $\pm$ 0.46 
& 12.69 $\pm$ 0.45 & 24.03 $\pm$ 1.88 \\
Graph WaveNet & 1296K & 12.37 $\pm$ 0.05 & 21.07 $\pm$ 0.16 
& 13.18 $\pm$ 0.22 & 23.00 $\pm$ 0.40 \\
DeepLGR & 968K & 13.82 $\pm$ 0.18 & 25.84 $\pm$ 0.45 
& 12.09 $\pm$ 0.06 & 21.48 $\pm$ 0.08 \\
\hline
PRNet (Ours) & 711K &\bf{11.76 $\pm$ 0.02} &  \bf{20.19 $\pm $ 0.04} 
& \bf{12.01 $\pm$ 0.02} & \bf{21.12 $\pm$ 0.05} \\
\hline
\hline
\multirow{2}{*}{\textbf{Method}} &
\multirow{2}{*}{\textbf{\# Params}}&
\multicolumn{2}{c|}{\textbf{P3}} & \multicolumn{2}{c}{\textbf{P4}}\\\cline{3-6}
&& MAE & RMSE  &  MAE & RMSE  \\
\hline
HA & - & 14.98 & 29.22 & 19.33 & 40.66 \\
DeepST & 380K & 14.94 $\pm$ 0.17 & 25.11  $\pm$ 0.14 
& 15.31 $\pm$ 0.35  & 27.45 $\pm$ 0.74\\
ST-ResNet& 3077K & 13.35 $\pm$ 0.10 & 23.36 $\pm$ 0.32 
& 13.39 $\pm$ 0.16 & 24.54 $\pm$ 0.02 \\
ConvLSTM& 1839K & 12.40 $\pm$ 0.03 & 22.12 $\pm$ 0.05 
& 12.07 $\pm$ 0.10  & 23.70 $\pm$ 0.23\\
DeepSTN+& 105M & 12.21 $\pm$ 0.02 & 21.89 $\pm$ 0.23
& 12.22 $\pm$ 0.11 & 24.15 $\pm$ 0.34 \\
Graph WaveNet & 1296K & 13.40 $\pm$ 0.16 & 23.98 $\pm$ 0.19
& 13.24 $\pm$ 0.21 & 25.58 $\pm$ 0.54\\
DeepLGR & 968K & 12.19 $\pm$ 0.06 & 21.91 $\pm$ 0.15
& 12.39 $\pm$ 0.14 &  24.09 $\pm$ 0.25 \\
\hline
PRNet (Ours) & 711K &\bf{ 12.09 $\pm$ 0.02}  & \bf{21.70 $\pm$ 0.04} & \bf{11.90 $\pm$ 0.05} & \bf{23.25 $\pm$ 0.13} \\
\hline
\end{tabular}
\end{table*}

More details of the two datasets are listed in Table~\ref{dataset_details}. We observe that most datasets have a high missing ratio. To address the missing ratio issue, there are two widely used strategies: Strategy 1 removes the segments with missing values~\cite{zhang2016dnn, zhang2017deep}, \revise{but significantly} reducing the number of samples; Strategy 2 fills the missing values with zero \revise{to produce} a larger data budget, while it introduces extremely noisy data. 
As the crowd flow shows periodic patterns, we take advantage of the periodicity to address this problem. Specifically, we use a dictionary to store a default periodic value for each time slot and use it to fill the missing value.
In the experiments, we set the default value as the mean of known values at the same time slot every week \revise{for an affordable way}. 
For example, the default value for Mon 7:00 is the \revise{average of the} known values on Mon 7:00. The missing value occurring in weekly segments (i.e., periodic closeness, and periodic prediction) will be filled with the default value. And samples with the missing value in the target segment for prediction \revise{are} discarded.
We employ the last 20\% data as the test set, and randomly select the remaining 60\% data as the training set and 20\% as the validation set, respectively. For a fair comparison, the same data preprocessing strategy is adopted for the models, including baselines models.

\subsubsection{Evaluation Metrics} 
Following the previous studies~\cite{zhang2017deep, lin2019deepstn+}, we evaluate our model using two metrics:
\textbf{Mean Absolute Error} (MAE) and \textbf{Root Mean Squared Errors} (RMSE).

\subsubsection{Implementation Details}
Our model is trained on a single GTX 2080 Ti using Adam optimizer with a learning rate of 0.0005. We set $T_{obs}$ to 12, $T_{pred}$ to 12, $D$ to 2, $C$ to 64, and $M$ to 9. The convolution kernel size in $\mathbf{W}_{f1}, \mathbf{W}_{f2}$ is 3 $\times$ 3 with 64 filters.
$H^{\prime}$ and $W^{\prime}$ are set to 8. $r_{s}$ and $r_{c}$ are 8 and 4, respectively. We apply a scalar with 50 on taxi volume. The early-stop strategy is applied in all the experiments. The maximum epoch is set to 250.

\subsubsection{Baselines}
We compare our model with seven baselines:
\begin{itemize}[leftmargin=*]
   
  \item  \textbf{HA} is a traditional time series method that averages the historical flow of the same time slot of the same day given \revise{past weekly segments}.
  
  \item \textbf{DeepST}~\cite{zhang2016dnn} is the first deep learning-based approach for grid-based crowd flow prediction, which utilizes convolution operators to extract local spatial correlations and different CNN branches to capture temporal dependencies.

  \item \textbf{ST-ResNet}~\cite{zhang2017deep} further enhances DeepST by introducing residual structure to improve the prediction accuracy.
  
  \item \textbf{ConvLSTM}~\cite{convlstm} integrates the convolution operation to RNN structure to enhance the long-term \revise{ST} relationship modeling.
  
  \item \textbf{DeepSTN}~\cite{lin2019deepstn+} \revise{uses} ordinary convolutions and fully-connected layers to capture the local and long-range spatial features, respectively.
  
  \item \textbf{Graph WaveNet}~\cite{graphwvenet} \revise{utilizes} graph neural network to learn self-adaptive spatial interaction and employ stacked dilated casual convolutions to capture long sequence dependency.
  
  \item \textbf{DeepLGR}~\cite{liang2020revisiting} adopts SE mechanisms to capture spatial correlation and temporal dynamics concurrently.
  
\end{itemize}
\begin{table}[!b]
\small
\vspace{-1em}
\tabcolsep=3.2mm
\centering
\caption{
Model comparison on BikeNYC dataset.
}
\label{quantitative_result_bikenyc}
\begin{tabular}{l|c|cc}
\hline
\textbf{Method} & \# Params & MAE &  RMSE   \\
\hline
HA & - & 3.38 & 7.52 \\
DeepST & 143K & 3.75 $\pm$ 0.06 & 7.50 $\pm$ 0.10 \\
ST-ResNet & 2841K & 3.60 $\pm$ 0.02 & 7.32 $\pm$ 0.03 \\
ConvLSTM & 1839K & 3.69 $\pm$ 0.07 & 8.20 $\pm$ 0.21  \\
DeepSTN & 1594K & 3.58 $\pm$ 0.05 & 7.72 $\pm$ 0.07 \\
Graph WaveNet & 1296K & 3.97 $\pm$ 0.04 &  8.20 $\pm$ 0.11\\
DeepLGR & 878K & 3.30 $\pm$ 0.03 & 7.57 $\pm$ 0.09 \\
\hline
PRNet (Ours) & 711K & \textbf{ 3.27 $\pm$ 0.01} & \textbf{7.08 $\pm$ 0.02}\\
\hline
\end{tabular}
\end{table}

\subsection{Experimental Results and Analysis}
Table~\ref{quantitative_result} and Table~\ref{quantitative_result_bikenyc} show the prediction results of baselines and our model on two datasets. 
The results \revise{show} that our model consistently outperforms existing methods on all datasets.
From the results, we can observe that: 
1) Traditional methods can outperform deep learning approaches \revise{on} some datasets, indicating that periodic information is an important characteristic for crowd flow prediction. 
For example, HA surpasses DeepST and ST-ResNet in P2.
The reason is that P2 has a 16.3\% missing ratio which causes the size of training samples to be relatively small so that \revise{deep models} become overfitted. 
\revise{However, HA achieves inferior performance under a larger data budget (e.g. P1, P3, and P4) as it is a nonparametric model, which ignores the ST correlation and the time trend.}
Our model takes advantage of traditional methods by integrating explicit periodic knowledge to guide the network, and therefore achieves the best performance among all methods across all datasets.
2) ConvLSTM, DeepSTN, Graph WaveNet, and DeepLGR show better results compared to DeepST and ST-ResNet, which demonstrates better \revise{ST} correlation understanding can lead to better performance.
We notice that DeepLGR can outperform Graph WaveNet, even though Graph WaveNet has a stronger temporal network, i.e., causal convolution network. We think this is because \revise{spatial correlations} are fully learned in CNNs via spatial kernels of each layer, while they are predefined in GNNs.
3) Our method achieves superior performance over Graph WaveNet, \revise{DeeSTN}, DeepLGR.
Specifically, it reduces MAE error by 8.49\%, 5.48 \%, 5.41 \% on average on the TaxiBJ dataset with 1.82, 147.68, and 1.36 times fewer parameters.
On the BikeNYC dataset, it also achieves competitive results with fewer parameters. 
\revise{Note} that the parameter numbers of DeepST, ST-ResNet, DeepSTN, and DeepLGR on TaxiBJ and BikeNYC are different.
Because their region-specific design for external feature encoding or spatial feature extraction leads to parameter growth as grid cells grow, especially for DeepSTN where fully-connect layers are used \revise{to capture} global spatial dependencies.
Surprisingly, ConvLSTM can produce favorable results. \revise{The reason could} be that the gate mechanism of it helps the model to capture better temporal dependencies in multi-step ahead prediction. However, it also requires high memory usage. 
Instead, our model utilizes a periodic residual learning strategy to provide stationary features to increase network capacity with fewer parameters. 
Overall, our work beats all the methods, \revise{proving that the network with a well-designed periodic learning strategy can make good crowd flow predictions.}

\begin{table}[ht]
\centering
\small
\vspace{-0.5em}
\tabcolsep=3.8mm
\caption{
Model w/ PRNet vs. w/o PRNet on TaxiBJ dataset.
}
\vspace{-0.5em}
\label{baselinewpr}
\begin{tabular}{l|c|cc}
\hline
\textbf{Method} & \textbf{\# Params}& \textbf{P1} & \textbf{P2} \\
\hline
DeepST & 380K & 15.68 $\pm$ 0.43 
& 15.61 $\pm$ 0.35 
\\
\rowcolor{background_gray} DeepST+ & 369K & 13.17 $\pm$ 0.14 
& 12.69 $\pm$ 0.09 
\\
ST-ResNet & 3077K & 13.84 $\pm$ 0.13
& 13.74 $\pm$ 0.42 
\\
\rowcolor{background_gray} ST-ResNet+ & 2503K & 11.95 $\pm$ 0.07
& 12.38 $\pm$ 0.03 
\\
DeepLGR & 968K & 13.82 $\pm$ 0.18
& 12.09 $\pm$ 0.06 
\\
\rowcolor{background_gray} DeepLGR+ & 893K 
& 11.78 $\pm$ 0.02
& 12.05 $\pm$ 0.06 
\\
\hline
\hline
\textbf{Method} & \textbf{\# Params}& \textbf{P3}& \textbf{P4}\\
\hline
DeepST & 380K 
& 14.94 $\pm$ 0.17 
& 15.31 $\pm$ 0.35 \\
\rowcolor{background_gray} DeepST+ & 369K 
& 12.71 $\pm$ 0.03 
& 13.88 $\pm$ 0.04 \\
ST-ResNet & 3077K 
& 13.35 $\pm$ 0.10 
& 13.39 $\pm$ 0.16 \\
\rowcolor{background_gray} ST-ResNet+ & 2503K 
& 12.32 $\pm$ 0.04 
& 12.21 $\pm$ 0.02  \\
DeepLGR & 968K 
& 12.19 $\pm$ 0.06 
& 12.39 $\pm$ 0.14  \\
\rowcolor{background_gray} DeepLGR+ & 893K
& 12.10 $\pm$ 0.06 
& 11.97 $\pm$ 0.02 \\
\hline
\end{tabular}
\vspace{-0.5em}
\end{table}

\subsection{Study on Periodic Residual Learning}
\label{sec:exp_res_learning}
We further verify the effectiveness of our periodic residual learning structure on different baselines in Table~\ref{baselinewpr}.
\revise{We use \textbf{DeepST+}, \textbf{ST-ResNet+} and \textbf{DeepLGR+} to denote the model using our proposed periodic residual learning structure with the ST network adopted from DeepST, ST-ResNet and DeepLGR, respectively.}
In other words, we adopt the backbone of DeepST, ST-ResNet, and DeepLGR as ST Module in PRNet. 
Also, we keep the EXT Module and the Decoder the same as their original networks.
As compared in Table~\ref{baselinewpr}, the networks with our proposed structure outperform their original models in terms of accuracy and robustness.
The proposed structure assists them to reduce the MAE error by 14.77\%, 10.52 \%, 5.13\%, and to promote the robustness (i.e., reduce the standard deviation) by 76.92\%, 80.25\%, 63.64\% on average \revise{on TaxiBJ dataset.}
The parameters of models are also reduced. Because the proposed structure encodes each observed segment with a shared ST Module rather than feeding them into different branches or concatenating them as a tensor.
By applying the shared \revise{network} to each time interval, PRNet provides explicit periodic references to the target segment from its corresponding periodic segments. Thus, it aids the model to increase the accuracy and robustness in long-term prediction, even with fewer parameters.
In summary, these results demonstrate the generality of our periodic residual learning structure across different networks.

\subsection{Ablation Study}
\label{sec:ablation}
Table~\ref{ablation_study} illustrates the effectiveness of each component in PRNet.
\textbf{SCE} adopts a single SCE Encoder to encode the closeness, periodic closeness, and periodic predictions, which is equivalent to the PRNet without the periodic residual learning mechanism.
\textbf{SCE w/o PC} only adopts a single SCE Encoder to encode the closeness and periodic predictions.
\textbf{SCE w/o S} is SCE model without the Spatial Enhance Module (SEM).
\textbf{w/o R} is PRNet without the residual learning module, which utilizes seven shared parameters SCE Encoders to encode seven observed segments.
\textbf{w abs} uses PRNet to predict the absolute values rather than residuals.

\begin{table}[h]
\small
\tabcolsep=3.4mm
\begin{center}
\caption{Ablation studies of PRNet on TaxiBJ-P4.
}
\label{ablation_study}
\begin{tabular}{l||c|c|c}
\hline
Method & \# Params & MAE & RMSE \\
\hline
SCE & 712K & 12.12 $\pm$ 0.11 & 23.78 $\pm$ 0.15\\ 
SCE w/o PC & 707K & 12.29 $\pm$ 0.16 & 24.34 $\pm$ 0.30\\ 
SCE w/o S & 703K & 12.39 $\pm$ 0.29  & 24.65 $\pm$ 0.97\\ 
w/o R & 703K &  36.61 $\pm$ 0.41 & 62.20 $\pm$ 0.47 \\ 
w abs & 711K & 12.31 $\pm$ 0.07 & 24.35 $\pm$ 0.35\\
PRNet (Ours) & 711K & \textbf{11.90 $\pm$ 0.05} & \textbf{23.25 $\pm$ 0.13}\\ 
\hline
\end{tabular}
\end{center}
\end{table}

According to the results shown in Table~\ref{ablation_study}, we can observe that: 1) Periodic closeness is important in prediction tasks. The reason is that it can provide the reference for time-series shifting between periodic predictions and future conditions.
2) Residual learning is essential for our model. Without it, the model cannot capture the correlations between the multi-scale time intervals because it only encodes the historical observations with seven shared parameters SCE Encoders separately.
Differing from \textbf{SCE w/o PC} and \textbf{SCE} that model the periodic pattern implicitly by fusing all observations into one SCE Encoder, PRNet directly calculates dependencies of multi-scale time intervals, which provides an elegant solution for explicit periodicity representation without introducing redundant parameters.
3) Enhancing spatial information boosts model performance. Our SEM provides the most salient features based on city-scale grids, which further promotes model performance.
4) Predicting the residual instead of absolute values leads to noticeable improvement, which proves our assumption about learning residual is much easier.
In summary, the experimental results and parameter comparison show that PRNet successfully captures the periodicity information as well as complex spatio-temporal correlations without increasing the model complexity.

\subsection{Effects of Hyperparameters}
\label{sec:hyper}

\begin{figure}[!htbp]
\vspace{-0.5em}
\begin{subfigure}{0.23\textwidth}
  \centering
    \includegraphics[height=3.4cm]{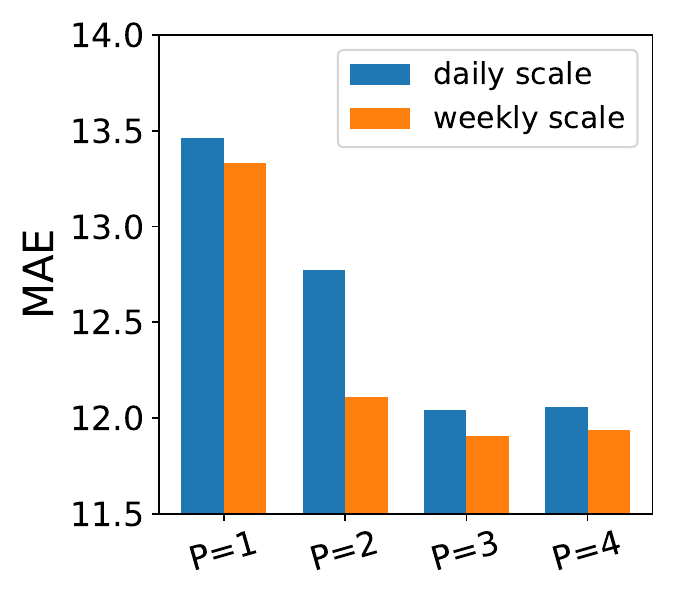}
  \caption{Scales with $P$ vs. MAE}
\end{subfigure}
\begin{subfigure}{0.23\textwidth}
  \centering
  \includegraphics[height=3.4cm]{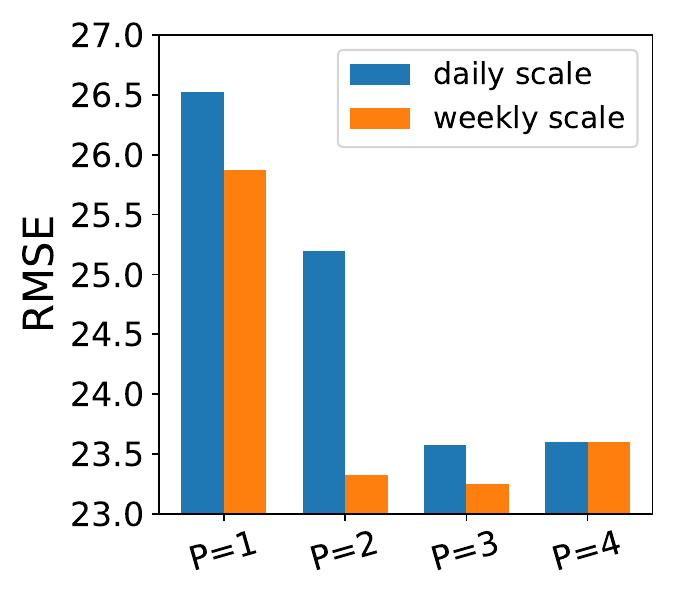} 
  \caption{Scales with $P$ vs. RMSE}
\end{subfigure}
\begin{subfigure}{0.23\textwidth}
  \centering
    \includegraphics[height=3.4cm]{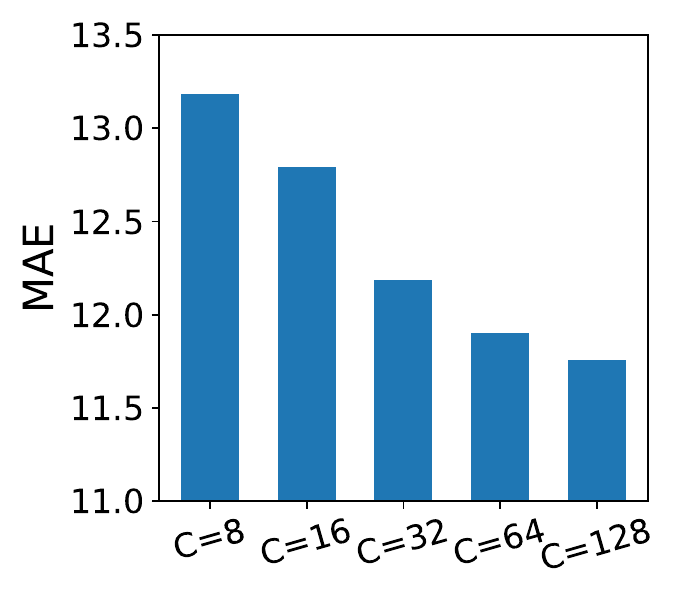}
  \caption{No. of channels vs. MAE}
\end{subfigure}
\begin{subfigure}{0.23\textwidth}
  \centering
  \includegraphics[height=3.4cm]{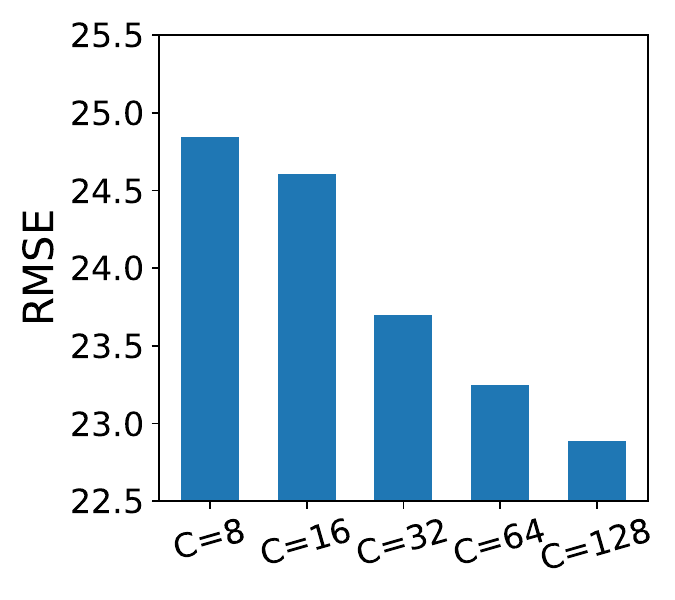} 
  \caption{No. of channels vs. RMSE}
\end{subfigure}
\vspace{-0.5em}
\caption{Effects of hyperparameters on TaxiBJ-P4.
}
\label{fig:periodicscale}
\vspace{-0.5em}
\end{figure}

In Fig.~\ref{fig:periodicscale}, we study the effects of hyperparameters in PRNet over TaxiBJ-P4.
First, we study the effects of different periodic scales \revise{with different} number of selected periods $P$ in Fig.~\ref{fig:periodicscale}(a)-(b).
Specifically, we explore different $P$ values from 1 to 4 under the daily scale and the weekly scale. \revise{The length of period $l$ is set to one day for daily scale and one week for weekly scale.}
From the results, we can observe that: 1) The network with the weekly scale consistently outperforms its daily scale counterpart. The reason is that weekly differencing can provide more stable residuals as the weekly crowd flow pattern tends to be similar, while the crowd flow pattern of two successive days can be different (e.g., Friday \revise{evening} and Thursday \revise{evening}). 
2) Increasing the number of $P$ can improve the model performance.
Because by considering multiple periodic segments, the network is allowed to attend to the information from different residual references. Then the network can perform more robustly \revise{even if} the sudden variation happens between two periodic segments.
However, using a very large $P$ value (i.e., $P = 4$) also \revise{degrades the performance} as the training samples are reduced.
We achieve the lowest MAE and RMSE when the $P$ value is 3, therefore, we choose $P=3$ as our default setting.
We also study the \revise{effects} of the number of channels \revise{$C$ by attempting different values of $C$ (i.e., 8, 16, 32, 64, and 128) in Fig.~\ref{fig:periodicscale}(c)-(d)}.
\revise{According to the results}, increasing $C$ from 8 to 128 can reduce the MAE and RMSE \revise{because} the model capability is improved. However, the network with 128 channels has 2802\revise{K} parameters, indicating it needs 3.94 times more parameters than the network with 64 channels. As our model achieves good performance when $C$ is 64, we use $C=64$ as \revise{the} default setting. 

\subsection{Study on Training Data Budget}
\label{sec:databudget}
\begin{figure}[!t]
\vspace{-0.5em}
\begin{subfigure}{0.23\textwidth}
  \centering
    \includegraphics[height=3.2cm]{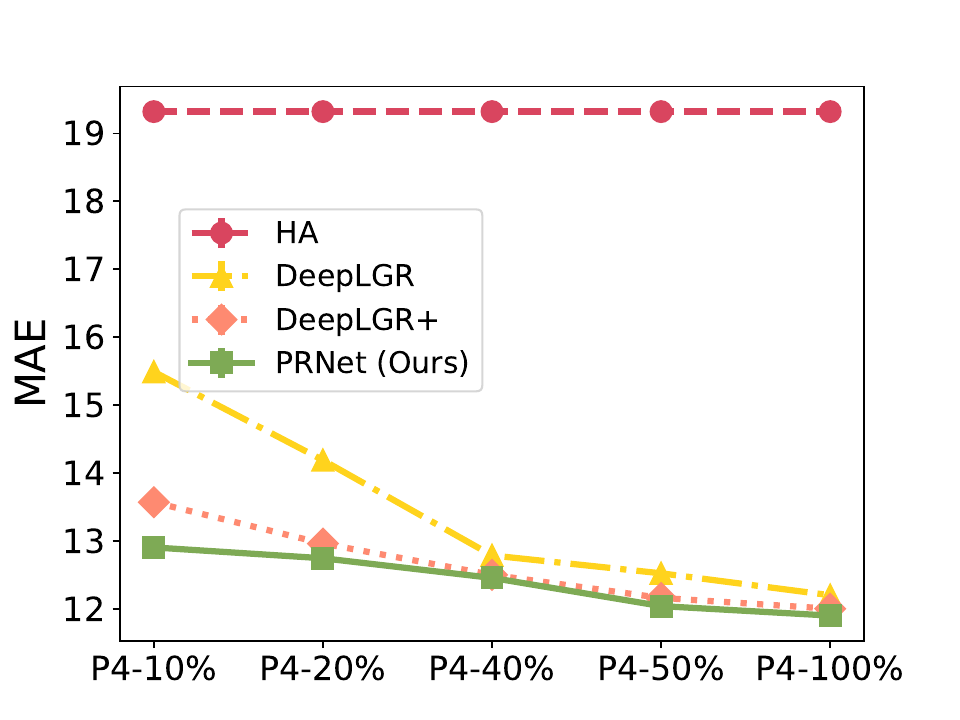}
  \caption{Result on MAE}
\end{subfigure}
\begin{subfigure}{0.23\textwidth}
  \centering
  \includegraphics[height=3.2cm]{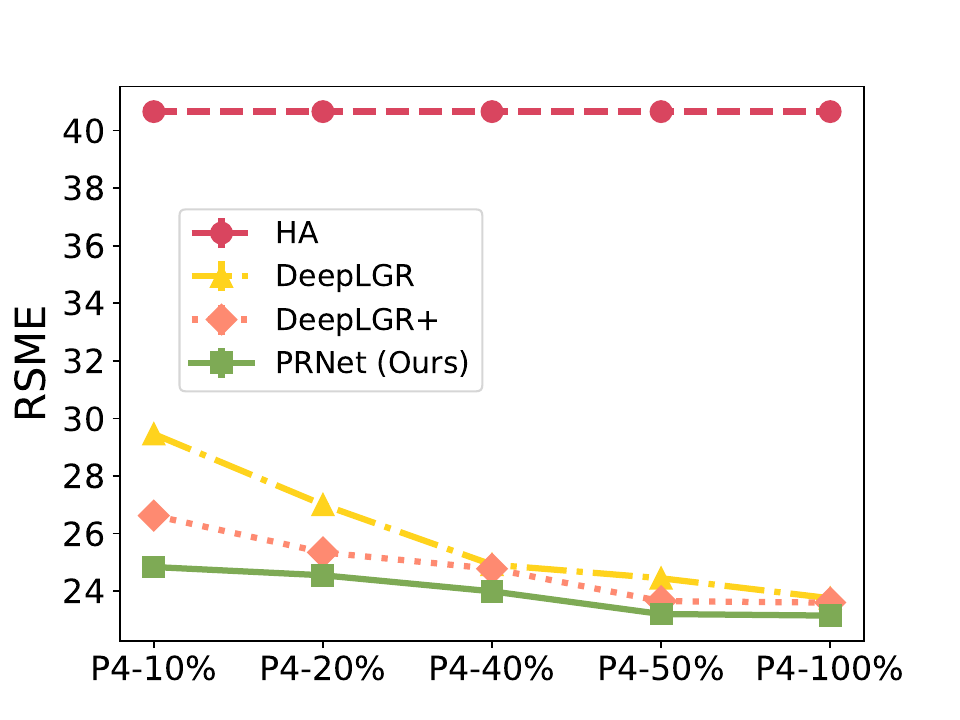} 
  \caption{Result on RMSE}
\end{subfigure}
\vspace{-0.5em}
\caption{Prediction results of PRNet under various data budgets. The training data are sampled from the original dataset with different ratios, i.e., 10\%, 20\%, 40\%, 50\%, and 100\%.}
\label{fig:ratio_train}
\end{figure}

In real-world applications, the available data budget for network training \revise{may varied}. Thus, we investigate the performance of our proposed network under different sizes of training data budgets \revise{on TaxiBJ-P4 in Fig.~\ref{fig:ratio_train}}. 
From the results, we can observe that the models with our proposed structure, i.e., DeepLGR+ and PRNet, surpass both HA and DeepLGR given various sizes of training data budgets. Specifically, they outperform DeepLGR by a large margin given a small size of training data (10\% ratio data budget).
Because our proposed structure explicitly captures the periodic residual which works as a strong periodic prior that provides the statistical knowledge to the deep learning network. 
With the deep model capturing the complex ST correlations, this periodicity prior knowledge can make our model bridge the gap between the traditional method and deep method, and thus help the model to generate good results, especially with small training data budgets.

\subsection{Study on Strategies for Missing Data}
\label{missingdata}
\begin{figure}[!htbp]
\vspace{-0.5em}
\begin{subfigure}{0.23\textwidth}
  \centering
    \includegraphics[height=3.4cm]{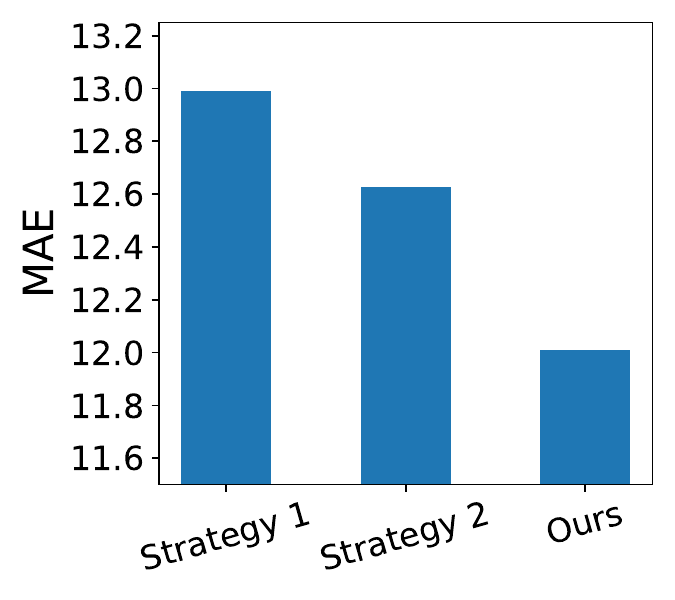}
  \caption{Result on MAE}
\end{subfigure}
\begin{subfigure}{0.23\textwidth}
  \centering
  \includegraphics[height=3.4cm]{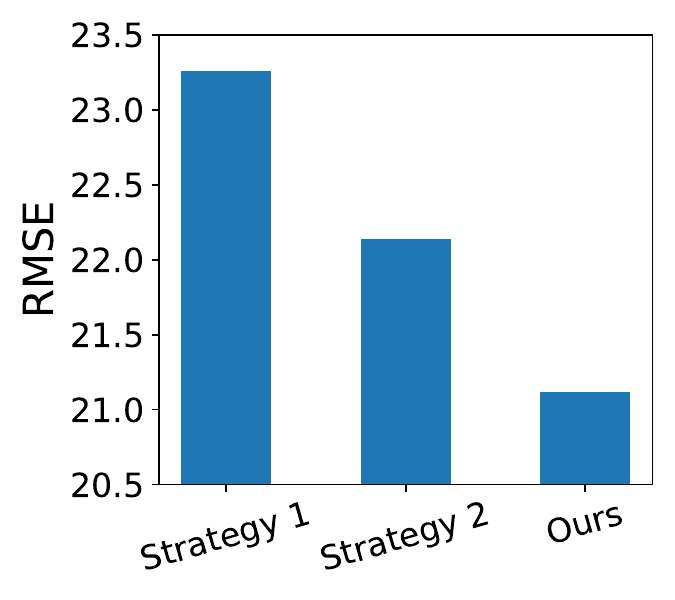} 
  \caption{Result on RMSE}
\end{subfigure}
\vspace{-0.5em}
\caption{Effect of missing data strategies on TaxiBJ-P2.}
\vspace{-0.5em}
\label{fig:missing_data_strategies}
\end{figure}

Fig.~\ref{fig:missing_data_strategies} shows the effectiveness of our proposed missing data strategy. We explore three strategies discussed in Section~\ref{sec:dataset} on TaxiBj-P4, which has a high missing ratio (i.e., 16.3\%). From the results, we can observe that: 1) Strategy 1 simply excludes the samples with missing data leading to a small data budget and therefore producing inferior performance. 2) Filling the missing data with periodic default values (\revise{Ours}) instead of replacing them with zero (\revise{Strategy 2}) boosts the model performance. Because it introduces less extremely noisy data.

\section{Conclusion and Future Work}
\label{sec:conclusion}
In this paper, we \revise{studied} the periodic behavior in crowd flow and proposed PRNet, a deep learning architecture that integrates the statistical strategy for multi-step ahead forecasting.
We further \revise{introduced} a lightweight SCE Encoder to enhance the spatio-temporal representation by suppressing and refining the intermediate features.
The experiments on real-world data \revise{shown} the effectiveness of PRNet, which reduces the error of MAE by 5.41\%$\sim$17.63\% with 1.36$\sim$147.7 times fewer parameters compared with SOTA methods.
Also, integrating PRNet into existing models \revise{reduced} the MAE error by 5.13\%$\sim$14.77\% and \revise{promoted} robustness by 63.64\%$\sim$80.25\%. 
It \revise{demonstrated} the potential of bridging the gap between the traditional time-series approaches and deep neural networks. 
This work highlights the inadequacy of previous works on periodicity modeling and sheds some light on exploiting traditional statistics to boost the deep learning model performance.
\revise{Moreover}, PRNet is not limited to crowd flow forecasting.
In the future, we will evaluate it on other tasks that contain strong periodicity. 

\section{Acknowledgement} 
This research is supported by the National Research Foundation, Singapore under its Industry Alignment Fund – Pre-positioning (IAF-PP) Funding Initiative. Any opinions, findings and conclusions or recommendations expressed in this material are those of the author(s) and do not reflect the views of National Research Foundation, Singapore.

\bibliographystyle{ACM-Reference-Format}
\bibliography{sample-base}

\end{document}